\documentclass{article}

\usepackage{PRIMEarxiv}

\usepackage[utf8]{inputenc} 
\usepackage[T1]{fontenc}    
\usepackage{hyperref}       
\usepackage{url}            
\usepackage{booktabs}       
\usepackage{amsfonts}       
\usepackage{nicefrac}       
\usepackage{microtype}      
\usepackage{lipsum}
\usepackage{fancyhdr}       
\usepackage{graphicx}       
\usepackage{amsmath,amsfonts,amsthm,bm}
\usepackage{algorithm}
\usepackage{algorithmic}
\usepackage{subcaption}
\usepackage{color,array}
\graphicspath{{media/}}     

\title{Active-learning-based non-intrusive model order reduction
}

\author{
  Qinyu Zhuang\\
  Technical University of Munich \\
  Munich \\
  \texttt{qinyu.zhuang@tum.de} \\
   \And
  Dirk Hartmann \\
  Siemens Industrial Software GmbH \\
  Munich \\
  \texttt{hartmann.dirk@siemens.com} \\
  \And
  Hans -J. Bungartz \\
  Technical University of Munich \\
  Munich \\
  \texttt{bungartz@tum.de} \\
  \And
  Juan M. Lorenzi \\
  Siemens AG \\
  Munich \\
  \texttt{juan.lorenzi@siemens.com} \\
}

\begin{document}
\maketitle

\begin{abstract}
The Model Order Reduction (MOR) technique can provide compact numerical models for fast simulation. Different from the intrusive MOR methods, the non-intrusive MOR does not require access to the Full Order Models (FOMs), especially system matrices. Since the non-intrusive MOR methods strongly rely on the snapshots of the FOMs, constructing good snapshot sets becomes crucial. In this work, we propose a new active learning approach with two novelties. A novel idea with our approach is the use of single-time step snapshots from the system states taken from an estimation of the reduced-state space. These states are selected using a greedy strategy supported by an error estimator based Gaussian Process Regression (GPR). Additionally, we introduce a use case-independent validation strategy based on Probably Approximately Correct (PAC) learning. In this work, we use Artificial Neural Networks (ANNs) to identify the Reduced Order Model (ROM), however the method could be similarly applied to other ROM identification methods. The performance of the whole workflow is tested by a 2-D thermal conduction and a 3-D vacuum furnace model. With little required user interaction and a training strategy independent to a specific use case, the proposed method offers a huge potential for industrial usage to create so-called executable Digital Twins (DTs).
\end{abstract}


\section{Introduction}
\label{sec:introduction}
Model Order Reduction \cite{antoulas2005approximation,volkwein2012pod,willcox2002balanced}, as a key component of the concept of DT \cite{rasheed2019digital,hartmann2018model}, is increasingly attracting attention from different research fields. The conventional ways of reduced modelling, such as Krylov Subspace Method \cite{heres2005robust}, Proper Orthogonal Decomposition (POD) method \cite{lu2019review} and Discrete Empirical Interpolation (POD-DEIM) method \cite{chaturantabut2010nonlinear}, already prove the power of MOR technique in several application fields. However, these methods require detailed knowledge of FOMs, including but not limited to the governing equation and system matrices. Such methods are known as intrusive reduction methods. The applicability of intrusive methods is limited to cases where such information is available, which is often not the case when dealing with industrial engineering simulation and commercial Finite Element Method (FEM) software. Recently, the development of so-called non-intrusive reduction has attracted much attention recently.

Non-intrusive MOR methods make use of the data generated with FOM to build surrogate models which can accurately reproduce the physical behavior encoded in the data. Currently, most of data-driven MOR methods are based on Machine Learning. Recently, we used Feed-forward Neural Network and Runge-Kutta Neural Network to construct the surrogate ROMs \cite{zhuang_lorenzi_bungartz_hartmann_2021}. Other network architectures such as Long-short-term-memory (LSTM) Neural Network \cite{mohan2018deep} and Recurrent Neural Network (RNN) \cite{wang2020recurrent, kani2017dr} have also been tested for the purpose of reduced modelling. Among them, a Feed-forward Neural Network has simple structure and sufficient approximating capability. Therefore, in this work, the ROMs are built with such networks.

The first concerning point of data-driven MOR techniques is collecting and storing the training data. To improve the accuracy of the ROMs, we always welcome more training data from the FOMs. However, in reality, computing the FOM to gain the training data is very time-consuming and too much training data could also reach the limits of the hardware, especially when looking at large 3D simulation models used productively in industry. Therefore, collecting the training data smartly becomes one of the focuses of the data-driven MOR techniques.

We orient our discussion by focusing on two aspects relevant to snapshot collection: data quality and data quantity. The data quality could describe many aspects, e.g., the distribution of the data, the noise in the data, etc.. Unlike other application, the data in the MOR field has less noise since it is usually obtained directly from the FOM solver. Therefore, the main focus will be improving the distribution of the training data. There are some investigations \cite{bastista2004a,noemi2022training} talking about how the distribution of the data could influence the machine learning. However, this issue is hardly discussed in the context of MOR.

The other concerning point is how much data should be collected. To minimize the amount of training data, the most straightforward solution is to use greedy algorithm. Greedy algorithm is widely used in the world of intrusive MOR. For example, in \cite{bui2008model}, the selection of the FOM snapshots is converted into an optimization problem and the algorithm is applied to a steady problem of a mechanical model. However, without the intrusiveness, the data-driven non-intrusive MOR technique are facing extra difficulties. The first difficulty is finding good stopping criteria. The stopping criteria usually need to validate the ROM and compute the accuracy. However, currently most validation \cite{perez2014nonintrusive, przekop2012alternative} is test-case-dependent,  which implies that the general performance of the ROM cannot be evaluated. To address this a load-independent metric is proposed for a certain mechanical problem in \cite{kuether2016validation}. The second difficulty is the absence of an error estimator which can adaptively decide the data increment strategy. Such an error estimator is available in an analytical form for the intrusive MOR methods but lacking in the non-intrusive case. 

In this paper, we suggest an alternative approach to tackle the two points above. For the first point, instead using only long FOM trajectories as the training snapshots, we propose to first estimate a reduced-state space and collect many one-step FOM trajectories starting from states randomly distributed in the reduced-state space. For the second point, we propose to use a validator based on PAC learning theory \cite{haussler1990probably} to decide when to stop the offline training. The validated ROMs are meaningful for the deployment in real industrial use cases where complex time-dependent inputs can be expected. The error estimator to speed up the convergence is constructed in a data-driven way \cite{paul2015adaptive, guo2018reduced}. Similarly to \cite{guo2018reduced}, we use a Kriging surrogate model \cite{krige1951statistical} as the error estimator to decide the data increment strategy.

The whole paper is structured as follows: in \autoref{sec:model_order_reduction}, MOR based on Neural Networks is introduced. In \autoref{sec:active_learning_for_model_order_reduction}, the key components and the whole workflow of the new active-learning-based MOR is described in details. To validate the new approach, two numerical tests are performed in \autoref{sec:numerical_examples}. Finally, some conclusions are given in \autoref{sec:conclusion}.

\section{ANN-based MOR}
\label{sec:model_order_reduction}
The workflow of MOR consists of two steps: constructing a low-dimensional space and identifying the ROM in the reduced space.

Without loss of generality, we can assume the FOM is governed by the equation
\begin{equation}
    \bm{\dot y}(t) = \bm{f}(\bm{y}, \bm{\mu})
\label{eq:general_fom}
\end{equation}
where $\bm{y} \in R^{N}$ is the state of the FOM and $N$ is normally very large. $\bm{f}$ is the right hand side (RHS) function configured by the parameter vector $\bm{\mu}=\begin{bmatrix}\mu_1& \mu_2& \dots& \mu_{N_{m}}\end{bmatrix} \in R^{N_{m}}$.

The first step of generating a ROM is to construct the reduced space by finding a low dimensional space defined by a projection matrix $\bm{V}\in\mathbb{R}^{N\times n}$, such that the reduced state $\bm{y}_r \in \mathbb{R}^{n}$ is given by projecting FOM state $\bm{y}$ onto the reduced space:
\begin{equation}
    \bm{y}_r = \bm{V^T} \bm{y}
\end{equation}
The second step is model identification, which means finding the governing equation in the reduced space, expected to have the form:
\begin{equation}
    \bm{\dot y_r}(t) = \bm{f_r}(\bm{y}_r; \bm{\mu})
\label{eq:general_rom}
\end{equation}
where $\bm{f_r}$ is the low-dimensional representation of $\bm{f}$.

Different methods and method combination can be used to achieve these two steps. However, to keep the non-intrusiveness of the proposed approach in this contribution, we will use data-driven methods for both. For constructing the reduced space, POD is employed, and ANN is used for identifying the ROM.

\subsection{POD}
\label{sec:pod}
POD is a very mature solution to construct the reduced space. Plenty of introductory literature for POD method is available, e.g., \cite{lu2019review}. Here we will just briefly introduce it. The main ingredient of the POD method is a series of observations on the FOM states, which are usually called snapshots:
\begin{equation}
    \bm{Y}=\begin{bmatrix}\bm{y}^{(1)}&\bm{y}^{(2)}&\dots&\bm{y}^{(N_s)}\end{bmatrix}
\end{equation}
where $N_s$ is the number of the available snapshots and $\bm{y}^{(i)}$ is the $i^\text{th}$ column in the snapshot matrix $\bm{Y}$.

Applying the Singular Value Decomposition (SVD) to $\bm{Y}$:
\begin{equation}
    \bm{Y} = \bm{V}\bm{\Sigma}\bm{U^T}
\end{equation}
The left singular vectors $\bm{V}\in R^{N\times k}$ can be used as the reduced space basis. This space will minimize the $L^2$ approximation error for the given snapshots. We can truncate $\bm{V}$ at its $n^{\text{th}}$ column. By doing this, the size of the reduced space ($n$) can be chosen freely. A big space can retain more information in the original full space, but it will make the ROM run slower during the online phase. 

\subsection{Conventional snapshot sampling}
To increase the diversity of the observed FOM states in the snapshots, usually a hyper cubic parameter space is predefined as:
\begin{equation}
    \mathcal{M} = \left[\mu_\text{min}^{(1)}, \mu_\text{max}^{(1)}\right] \times \left[\mu_\text{min}^{(2)}, \mu_\text{max}^{(2)}\right] \times \dots \times \left[\mu_\text{min}^{(N_m)}, \mu_\text{max}^{(N_m)}\right]
\end{equation}
where $\mu^{(i)}$ is the $i^\text{th}$ component of the system parameter vector $\bm{\mu}$. The parameter vectors $\bm{\mu}$ will be randomly selected from $\mathcal{M}$ and different initial value problems (IVPs) can be constructed based them. In this case, FOM's states under different input parameters can be observed.

Here we introduce two ways called Static Parameter Sampling (SPS) and Dynamic Parameter Sampling, respectively. For SPS, to construct $m$ IVPs, $m$ parameter vectors will be sampled from $\mathcal{M}$. For IVP $i$, we have:
\begin{equation}
    \bm{\dot y}(t) = \bm{f}(\bm{y}, \bm{\mu}^{(i)})
\end{equation}
where $\bm{\mu}^{(i)}$ is the $i^\text{th}$ parameter samples.

With DPS, to construct $m$ IVPs where each IVP has $K$ time steps, $mK$ parameter vectors will be sampled from $\mathcal{M}$. For IVP $i$, we have:
\begin{equation}
    \bm{\dot y}(t) = \bm{f}(\bm{y}, \bm{\mu}^{(i)}(t))
\end{equation}
where:
\begin{equation}
    \bm{\mu}^{(i)}(t) = \text{Interp}(\{\bm{\mu}^{(i0)}, \bm{\mu}^{(i1)}, \dots, \bm{\mu}^{(iK)}\},\{t_0, t_1, \dots, t_K\})
\label{eq:interpolation}
\end{equation}
In \autoref{eq:interpolation}, $\text{Interp}(\cdot)$ means create a function by interpolating using two given sets. Let $\delta t$ be the time step size, $t_i=i\delta t$.

As we can find that DPS is $K$ times more efficient in exploring the parameter space $\mathcal{M}$, it is used to build the FOM with complex parameter configuration.

\subsection{ROM identification using ANNs}
\label{sec:ROM_identification}
The ROM we employ in this paper is a dynamic model evolving on a discrete time grid from $t_0$ to $t_{end}$:
\begin{equation}
    \frac{\bm{y}_{r,i+1}-\bm{y}_{r,i}}{\delta t} = \bm{g}(\bm{y}_{r,i}; \bm{\mu}_i)
\label{eq:rom_eq}
\end{equation}
where $\bm{y}_{r,i}$ means $\bm{y_{r}}(t=t_i)$. And $\bm{g}(\cdot)$ is the approximation to the unknown reduced RHS $\bm{f_r}(\cdot)$.

To learn the mapping between states of two consecutive time steps, a Multi-layer Perceptron (MLP) is used as $\bm{g}(\cdot)$ in \autoref{eq:rom_eq}. In this case the inputs to the Neural Network are the current state of the system $\bm{y}_{r,i}$ and $\bm{\mu}_i$, and the new reduced state $\bm{y}_{r,i+1}$ is calculated as
\begin{equation}
    \bm{y}_{r,i+1} = \bm{y}_{r,i} + \delta t \bm{g}_{\text{ee}}(\bm{y}_{r,i}, \bm{\mu}_i)
    \label{eq:eenn_rhs}
\end{equation}
this looks similar to an Explicit Euler integrator, so we denote the network's function as $\bm{g}_{\text{ee}}(\bm{y}_{r,i}, \bm{\mu}_i)$. And we call such a network Explicit Euler Neural Network (EENN) \cite{zhuang_lorenzi_bungartz_hartmann_2021, pan2018long}. From \autoref{eq:eenn_rhs}, we know that an EENN is essentially a variant of residual network \cite{he2016deep} which learns the increment between two system states instead of learning to map the new state directly from the old state. This leads to a potential advantage of EENNs that we can use deeper network for approximating more complex nonlinearity.

\begin{figure}[!htbp]
  \centering
  \includegraphics[width=0.7\textwidth]{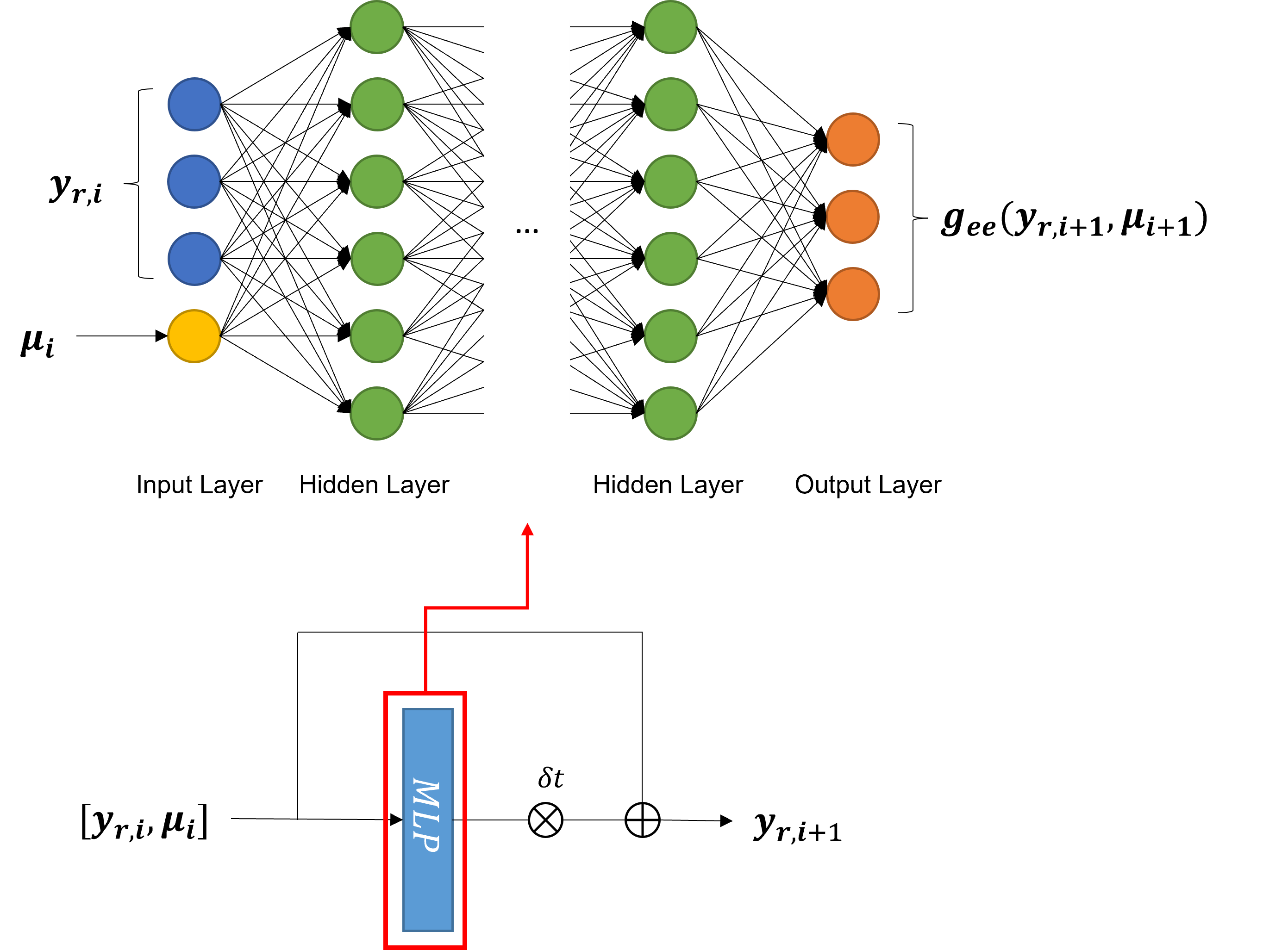}
  \caption{Sketch of an EENN. The EENN uses an MLP to approximate the RHS of \autoref{eq:rom_eq}. The output of the MLP is assembled as described in \autoref{eq:eenn_rhs}}
  \label{fig:eenn}
\end{figure}

The algorithm for training such an EENN is provided in \autoref{algo:EENN}  where $\bm{\hat{y}}_{r,i+1}$ means the reference solution and a sketch of the network structure is given in \autoref{fig:eenn}.

\begin{algorithm}                     
\caption{Training of EENN}          
\label{algo:EENN}                          
\begin{algorithmic} [1]                   
\REQUIRE $\bm{y}_{r,i}, \bm{\mu}_i, \bm{y}_{r,i+1}$
\WHILE{$\text{loss}>\text{loss}_{\text{tol}}$} 
\STATE{
$\bm{y}_{r,i+1} = \bm{y}_{r,i} + \delta t \bm{g}_{\text{ee}}(\bm{y}_{r,i}, \bm{\mu}_i)$
}
\STATE{
$\text{loss}=\text{MSE}(\bm{\hat{y}}_{r,i+1}, \bm{y}_{r,i+1})$
}
\STATE{
Use backpropagation to update the weights and biases of the network $\bm{g}_{\text{ee}}$
}
\ENDWHILE
\end{algorithmic}
\end{algorithm}

The EENN's core function $\bm{g}_{\text{ee}}(\bm{y}_{r,i}, \bm{\mu}_i)$ is an MLP consisting of one input layer, one output layer and multiple hidden layers. The residual structure allows the MLP to have more hidden layers but without suffering too much from network degeneration.

In this contribution, all networks consist of 1 input layer, 2 hidden layers and 1 output layer. The activation function between each 2 consecutive layers is Rectified Linear Unit (ReLU) function \cite{glorot2011deep}. We use Pytorch \cite{paszke2017automatic} as the platform to build and training the networks. The learning rate decay and Early Stopping \cite{prechelt1998early} is applied to facilitate the training.

\section{Active learning for MOR}
\label{sec:active_learning_for_model_order_reduction}
\begin{figure}[!htbp]
  \centering
  \includegraphics[width=0.6\textwidth]{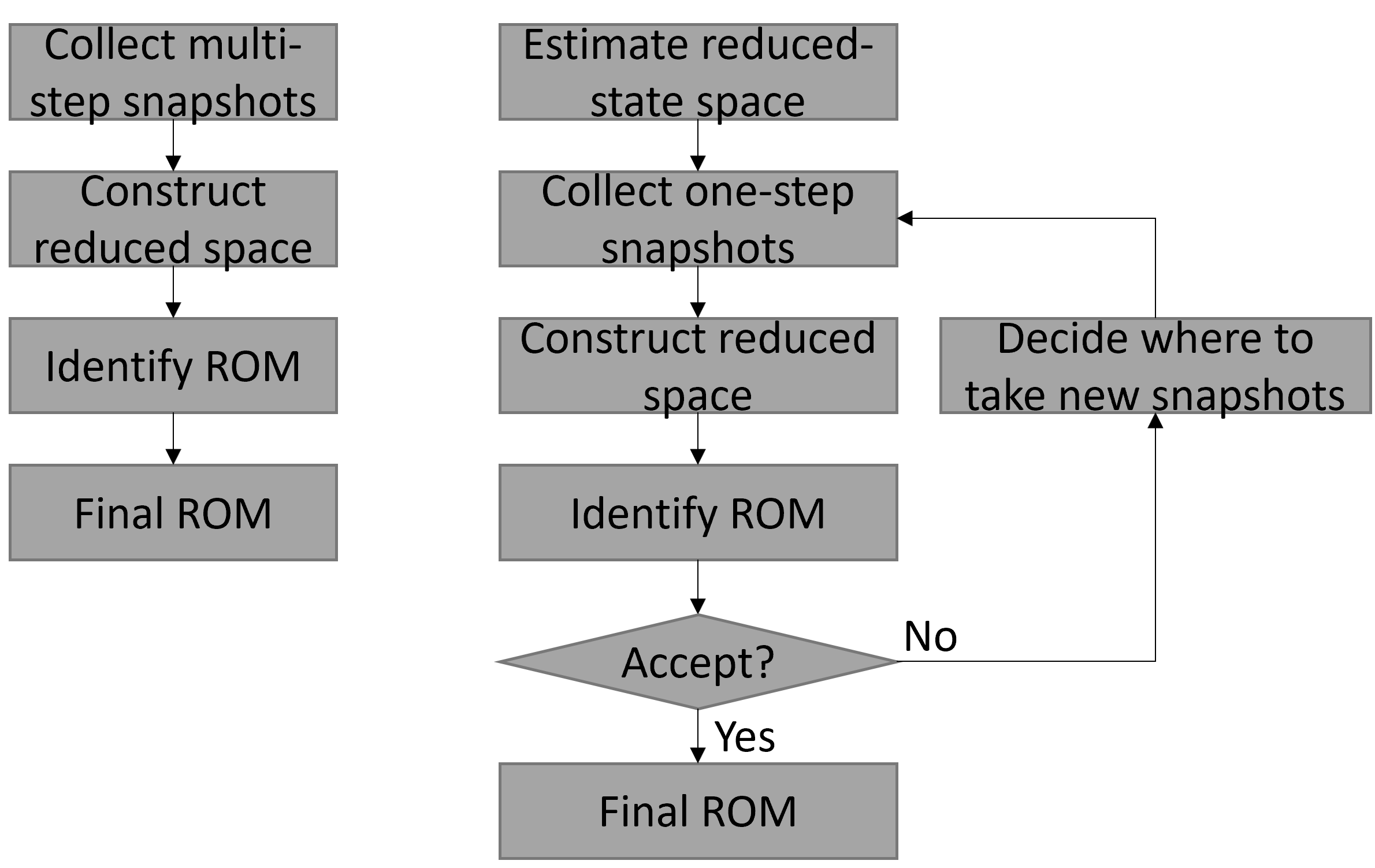}
  \caption{Workflow of machine-learning-based MOR (left) and the active-learning-based MOR (right)}
  \label{fig:algorithm_diagram}
\end{figure}
For most conventional non-intrusive MOR approaches, snapshots are collected up-front and used as a static input to the MOR process. This could cause the problem that not all important information is captured in the snapshots for constructing the ROM. Inspired by greedy snapshot collection in intrusive MOR \cite{haasdonk2011training, lappano2016greedy}, in this contribution, we propose a solution based on active-machine-learning which enables collecting snapshots dynamically without breaking the non-intrusiveness.

As we see in \autoref{fig:algorithm_diagram}, different from the conventional workflow, in the proposed workflow, we start with estimating reduced-state space. Instead of using long FOM trajectories results to construct the dataset, many one-step snapshots are collected from the space. They are used to train and validate the ROM. The validator is built based on the PAC learning and an error estimator based on GPR is employed to ensure the convergence.

\subsection{Reduced-state space estimation and one-step sampling}
\label{sec:one_step_snapshots}
In this contribution, we do not directly use long trajectories as the training data for MOR. Instead, reduced-state space will first be estimated with a few long trajectories. Sequentially, many one-step snapshots will be collected in such space and used as the training data. With the pre-defined parameter space $\mathcal{M}$, the boundaries of the reduced-state space will be estimated as follows:
\begin{enumerate}
    \item Use SPS to create $m$ IVPs for the FOM. Each IVP is from time $t_0$ to time $t_{\text{end}}$ with $K$ time steps.
    \item Solving all the FOM problems will produce the snapshot matrix $\bm{Y}_\text{estimate}=\begin{bmatrix}\bm{y}^{(1)}&\bm{y}^{(2)}&\dots&\bm{y}^{(m\times K)}\end{bmatrix}$ which contains $m\times K$ FOM state vectors.
    \item Apply POD on $\bm{Y}_\text{estimate}$ to construct the reduced space. And project $\bm{Y}_\text{estimate}$ onto the reduced space to get its low-dimensional representation:
    \begin{equation}
        \bm{Y}_{r, \text{estimate}} = \bm{V}^T \bm{Y}_\text{estimate}
    \end{equation}
    \item Find the minimum and the maximum entries for each POD component in $\bm{Y}_\text{estimate}$ and denote them as:
    \begin{equation}
        \left\{y_{r,\text{min}}^{(1)}, y_{r,\text{max}}^{(1)},y_{r,\text{min}}^{(2)}, y_{r,\text{max}}^{(2)},...,y_{r,\text{min}}^{(N_r)}, y_{r,\text{max}}^{(N_r)}\right\}
    \label{eq:yr_limits}
    \end{equation}
    where $N_r$ is the size of the ROM.
    \item The estimated space $\mathcal{Y}$ is:
    \begin{equation}
        \mathcal{Y}=\left[y_{r,\text{min}}^{(1)}, y_{r,\text{max}}^{(1)}\right]\times\left[y_{r,\text{min}}^{(2)}, y_{r,\text{max}}^{(2)}\right]\times\dots\times\left[y_{r,\text{min}}^{(N_r)}, y_{r,\text{max}}^{(N_r)}\right]
        \label{eq:estimated_Y}
    \end{equation}
\end{enumerate}

The number of simulations ($m$) can be selected as any positive integer. Since the reduced basis $\bm{V}$ computed in the step will be used as the initial reduced basis later in the greedy algorithm, the more simulations we have, the better the initial reduced basis will be. And while configuring the FOM simulation, we use constant system parameters, which means in $j^\text{th}$ simulation, $\bm{\mu}_i\equiv\bm{\mu}^{(j)}, i=0,1,...,K$. This makes the FOM evolve to the extreme states easier and the boundary values for the POD coefficients are most likely to be observed when the FOM is at the extreme states. Similarly, one can use minimal intrusiveness to the full order problem to improve the estimate. To be specific, the parameter configurations which most likely drive the system to its extreme status should be included into $\mathcal{M}$.

A sample $\bm{J}_\text{sample}$ from the joint space $\mathcal{J}=\mathcal{Y}\times\mathcal{M}$ has the form of $\bm{J}_\text{sample} = \begin{bmatrix}\bm{y}_{r,\text{sample}}&\bm{\mu}_\text{sample}\end{bmatrix}$. We can use $\bm{J}$, $\bm{V}$ and the FOM solver to perform a one-step simulation whose initial condition is $\bm{y}_{r,\text{sample}}$ and step size is $\delta t$. The simulation result will form a short (one-step) trajectory in $\mathcal{Y}$. The dataset constructed by such short trajectories will be used for training and validating the ROM in this paper.

\begin{figure}[!htbp]
    \centering
    \includegraphics[width=0.5\textwidth]{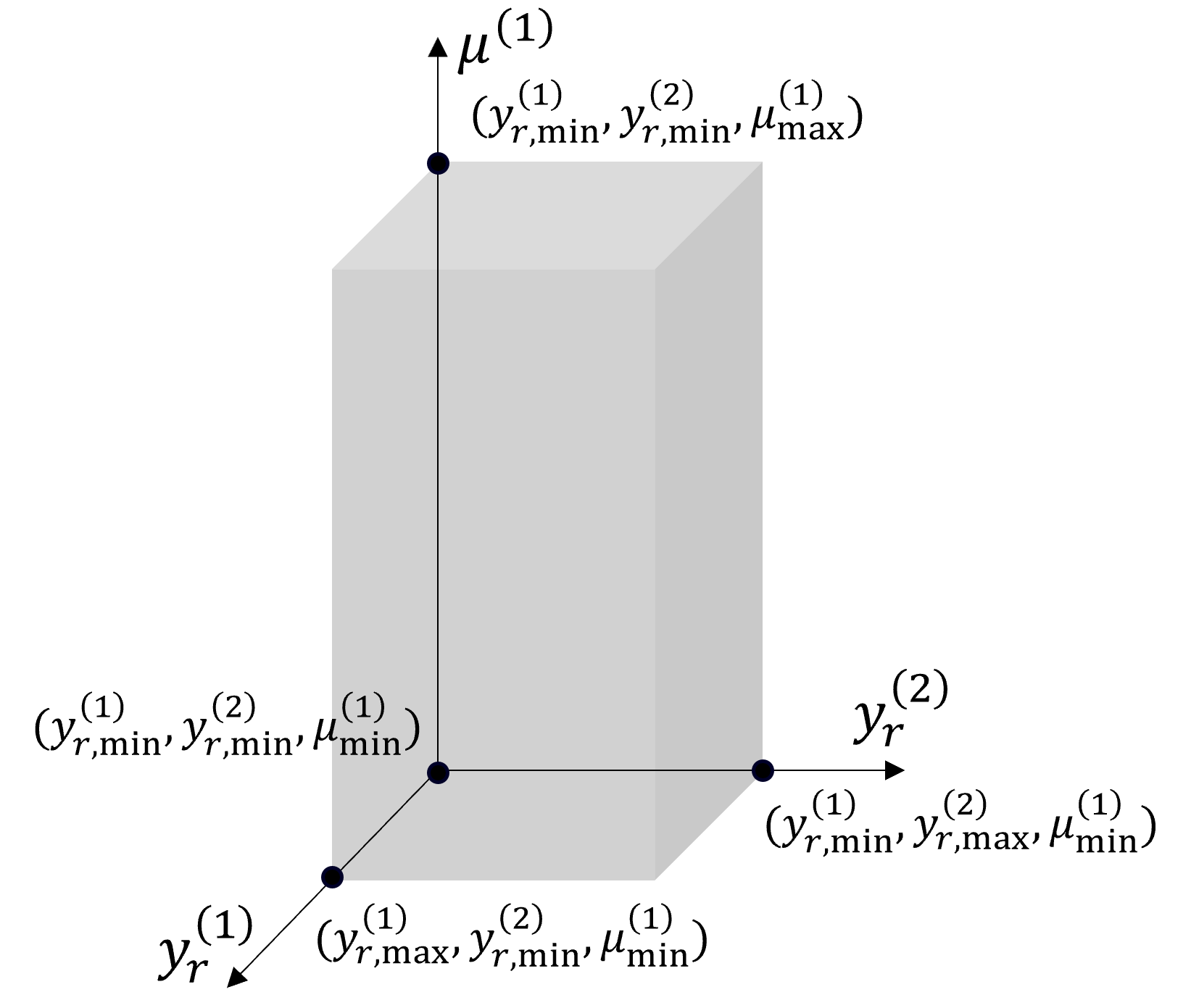}
    \caption{An estimated joint space $\mathcal{J}$ where $N_r=2$ and $N_u=1$.}
    \label{fig:J}
\end{figure}

\subsection{Loosen and trim reduced-state space}
\label{sec:loosen_and_trim_reduced_state_space}
\begin{figure}[!htbp]
    \centering
    \includegraphics[width=0.6\textwidth]{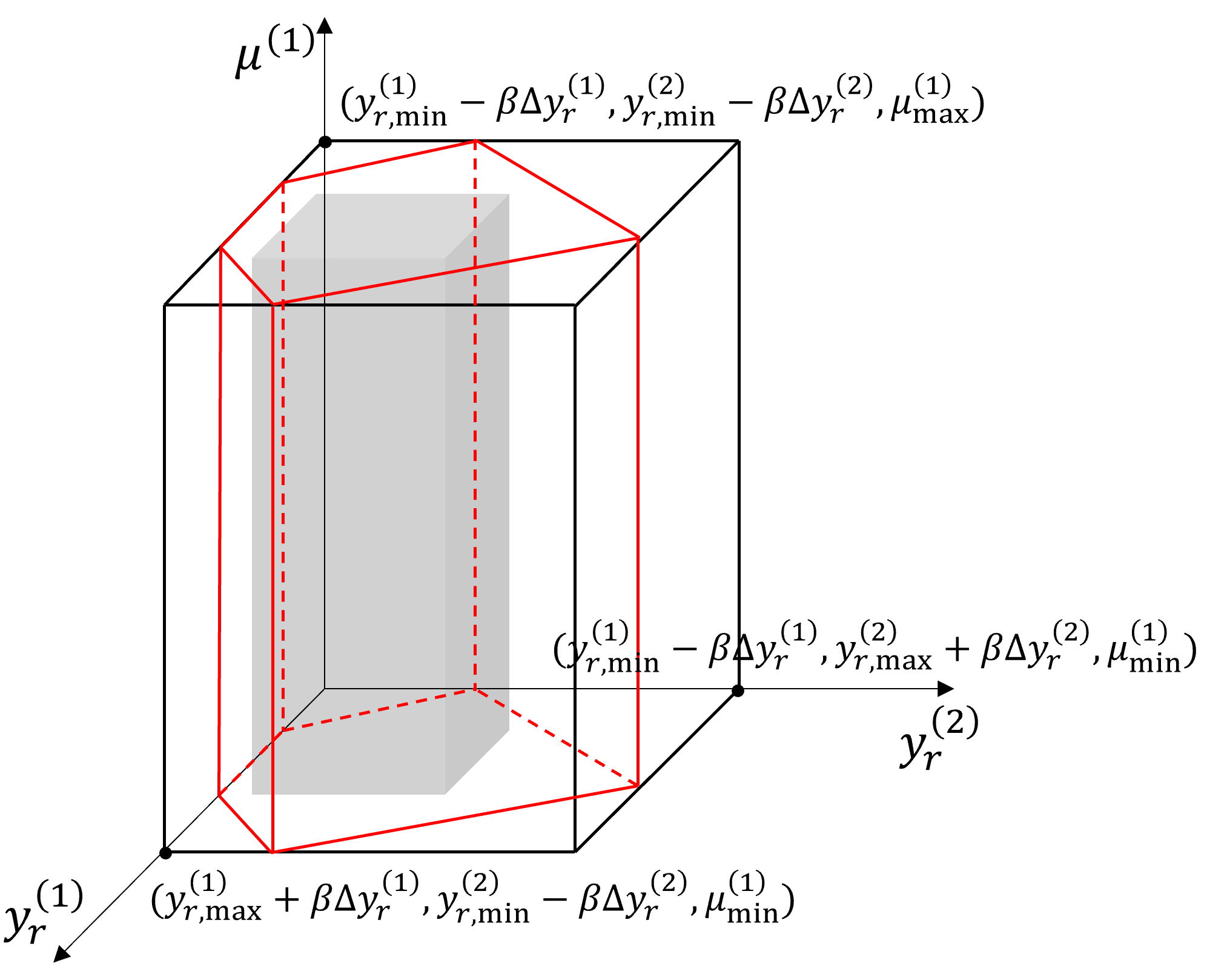}
    \caption{The joint space $\mathcal{J}$ in \autoref{fig:J} being loosened (black solid lines) and trimmed (red solid/dashed lines).}
    \label{fig:J_star}
\end{figure}

However, a joint space $\mathcal{J}$ created in \autoref{sec:one_step_snapshots} is a purely hyper-cubic space (see example in \autoref{fig:J}) which can still be improved. We will loosen and trim $\mathcal{Y}$ to make $\mathcal{J}$ more reasonable. The first step is loosening. And for this purpose, we introduce an expansion ratio $\beta$:
\begin{align}
    \begin{split}
        &\Delta y_{r}^{(i)} =  y_{r,\text{max}}^{(i)} -  y_{r,\text{min}}^{(i)}\\
        &\text{replace }y_{r,\text{min}}^{(i)}\text{ in \autoref{eq:estimated_Y} with }y_{r,\text{min}}^{(i)}-\beta\Delta y_{r}^{(i)}\\
        &\text{replace }y_{r,\text{max}}^{(i)}\text{ in \autoref{eq:estimated_Y} with }y_{r,\text{max}}^{(i)}+\beta\Delta y_{r}^{(i)}\\
    \end{split}
    \label{eq:loosened_Y}
\end{align}
To trim the space, we apply two conditions to a reduced state $\bm{y}_r$ sampled from $\mathcal{Y}$:
\begin{align}
    \begin{split}
        &\text{MAX}(\bm{V}\bm{y}_r)<y_\text{max}\\
        &\text{MIN}(\bm{V}\bm{y}_r)>y_\text{min}\\
    \end{split}
    \label{eq:trimming_condition}
\end{align}
where $\text{MAX}(\cdot)$ and $\text{MIN}(\cdot)$ means the maximum and minimum entry of a vector/matrix. The value $y_\text{min}$ and $y_\text{max}$ can be either defined by empirical values of the physics quantity $y$ or by $\text{MAX}(\bm{Y}_\text{estimate})$ and $\text{MIN}(\bm{Y}_\text{estimate})$. In this paper, $\text{MAX}(\bm{Y}_\text{estimate})$ and $\text{MIN}(\bm{Y}_\text{estimate})$ is used. We here denote the final feasible sampling space for reduced states as $\mathcal{Y}^*$, and the final joint space as $\mathcal{J}^*=\mathcal{Y}^*\times\mathcal{M}$. A graphical example for joint space $\mathcal{J}^*$ is given in \autoref{fig:J_star}.

\subsection{ROM validation}
\label{sec:ROM_validator}
Finding an adequate metrics to validate a ROM is challenging. When the training and test data sets are collected up front the ROM could be biased towards the data set and lead to lack of generalization for the model. To make our data-driven ROM more reliable, we need to design a more generalized validating procedure.

\paragraph{Accuracy} Here we consider using the ROM to predict the system states from time $t_0$ to time $t_{\text{end}}$ with time step $\delta t$. If there are $K$ steps stored in the prediction, we will have a time-series prediction as:
\begin{equation}
    \bm{Y}=\begin{bmatrix}\bm{y}_1&\bm{y}_2&\dots& \bm{y}_K\end{bmatrix}
\end{equation}
where $\bm{y}_i=\bm{y}(t=t_0+i\delta t)$. And with the solution $\hat{\bm{Y}}=\begin{bmatrix}\bm{\hat{y}}_1&\bm{\hat{y}}_2&\dots&\bm{\hat{y}}_K\end{bmatrix}$ computed by the FOM solver, we calculate the ROM relative error as:
\begin{equation}
    E = \text{Mean}\left(\frac{\left\|\bm{Y}-\hat{\bm{Y}}\right\|}{\left\|\hat{\bm{Y}}\right\|}\right)
\end{equation}
Here, the relative error deciding the accuracy of the ROM is calculated from a time-series prediction. A disadvantage of such method is the existence of the accumulative error. In such measurement, if the ROM makes mistakes on predicting for the first several time points, the errors will be accumulated to the following prediction and later error values become meaningless because of the wrong initial state. Therefore, we propose to use several independent one-step prediction for validating the ROM. However, the ROM is very likely to be used for a long-term prediction and such a one-step-prediction accuracy still cannot reflect the performance of the ROM. Therefore, we will introduce the confidence of the ROM.

\paragraph{Confidence} We define the relative error of the ROM in one-step-forward prediction as $e$:
\begin{equation}
    e = \frac{ \left\|\bm{y}-\bm{\hat{y}} \right\| }{ \left\|\bm{\hat{y}}\right\| }
\end{equation}
We furthermore define ROM's accuracy $1-\tau$ and its corresponding confidence $\eta$ to construct such an inequality:
\begin{equation}
    \text{Pr}(e\leq\tau)>\eta
\label{eq:error_ineq}
\end{equation}
i.e. the probability that the error $e$ smaller than $\tau$ is greater than $\eta$, which also means that our ROM is $\eta$ confident to having error lower than $\tau$. $\eta$ and $1-\tau$ is a pair of confidence and accuracy that we are seeking.

We assume we have $s$ independent tests in the validator. Then we can calculate the observed confidence $p(\tau^*)$ of a given accuracy $1-\tau^*$ using:
\begin{align}
\begin{split}
    p(\tau^*) = \frac{1}{s}\sum_{i=1}^s L(e_i\leq\tau^*)\\
    L(e_i\leq\tau^*) = 
    \begin{cases}
    1 & e_i\leq\tau^*\\
    0 & e_i>\tau^*\\
    \end{cases}
\end{split}
\label{eq:observed_confidence}
\end{align}
However, we can only include a limited number of independent tests into the validator. This means, with a given $\tau^*$, the $p(\tau^*)$ computed from the validator should also have an accuracy $\epsilon$ and confidence $1-\sigma$ from the true confidence $\hat{p}(\tau^*)$. Thus, we obtain another inequality:
\begin{equation}
    \text{Pr}(|p(\tau^*)-\hat{p}(\tau^*)|<\epsilon)>1-\sigma
\label{eq:observed_and_true_confidence}
\end{equation}
where $\hat{p}(\tau^*)$ is the true confidence. If we predefine the accuracy $1-\epsilon$ and the confidence $1-\sigma$ of the validator, the smallest number of the independent tests needed by the validator can be computed using Chernoff's inequality \cite{alippi2016intelligence}:
\begin{equation}
    s\geq s_{\text{PAC}} = \frac{1}{2\epsilon^2}ln\left(\frac{1}{\sigma}\right)
\label{eq:S_PAC}
\end{equation}
Let $\Bar{\tau}^*$ be the smallest $\tau^*$ satisfying:
\begin{equation}
    p(\Bar{\tau}^*) = 1
\label{eq:gamma_eq_1}
\end{equation}
Since Inequality \autoref{eq:observed_and_true_confidence} holds for all $\tau^*$, it also holds for $\Bar{\tau}^*$:
\begin{equation}
    \text{Pr}(|1-\hat{p}(\Bar{\tau}^*)|<\epsilon)>1-\sigma
\end{equation}
That is the inequality:
\begin{equation}
    |1-\hat{p}(\Bar{\tau}^*)|<\epsilon
\end{equation}
holds with the probability $1-\sigma$. And further we get:
\begin{equation}
    1-\epsilon<\hat{p}(\Bar{\tau}^*)<1
    \label{eq:rom_true_conf}
\end{equation}
holds with probability $1-\sigma$.

Finally, the ROM's accuracy $1-\tau$ and confidence $\eta$ are therefore $1-\Bar{\tau}^*$ and $1-\epsilon$ respectively. For other $\tau^*<\Bar{\tau}^*$, the ROM does not have the confidence $\eta$ of being $1-\tau^*$ accurate and $p(\tau^*)$ will be smaller than 1. In other words, the ROM cannot pass all tests in the validator with the accuracy $1-\tau^*$. Since Inequality \autoref{eq:rom_true_conf} holds with probability $1-\sigma$ and $S_\text{PAC}$ is negatively correlative to $\sigma$, we know that the more test samples we have, the more reliable the computed true confidence will be.

Using the method described in \autoref{sec:one_step_snapshots}, $s$ independent joint samples can be selected from the joint space $\mathcal{J}$. Given with the samples, the FOM solver will generate $s$ outputs. They are essentially $s$ system states evolved from the given initial states after $\delta t$ and they will be used as the reference outputs of the PAC validator. We will give the same inputs to the ROM and compare the ROM's output with the reference. Based on the comparison we can compute how accurate the ROM could be with the confidence of $\eta_{\text{design}}$. We denote such a validator as:
\begin{equation}
    p(\tau^*) = \text{PAC}(\tau^*;\text{ROM})
\end{equation}

As mentioned before, the best accuracy at the confidence level $\eta_{\text{design}}$ can be found by finding the smallest $\tau^*=\Bar{\tau}^*$ satisfying $p(\Bar{\tau}^*)=1$. This value will be another ingredient used for checking the convergence of the ROM.

\subsection{Error estimator}
\label{sec:error_estimator}
As sketched in \autoref{fig:algorithm_diagram}, the current ROM will go through all tests in the validator. We denote the ROM error on the tests as:
\begin{align}
\begin{split}
    &\bm{e} = \begin{bmatrix}e^{(1)}&e^{(2)}&\dots&e^{(s)}\end{bmatrix}\\
    &e^{(i)} = \frac{\left\|\bm{V}\text{ROM}(\bm{J}^{(i)})-\bm{\hat{y}}_\text{out}^{(i)}\right\|}{\left\|\bm{\hat{y}}_\text{out}^{(i)}\right\|}=\frac{\left\|\bm{y}_\text{out}^{(i)}-\bm{\hat{y}}_\text{out}^{(i)}\right\|}{\left\|\bm{\hat{y}}_\text{out}^{(i)}\right\|}
\end{split}
\end{align}
where $\bm{y}_\text{out}^{(i)}\text{ and }\bm{\hat{y}}_\text{out}^{(i)}$ is the predicted and the reference output corresponding to the input $\bm{J}^{(i)}=[\bm{y}_r^{(i)}, \bm{\mu}^{(i)}]$ after one time step, respectively.
We assume there is a function:
\begin{equation}
    e^{(i)} = \mathcal{E}(\bm{J}^{(i)})
\label{eq:error_interpolator}
\end{equation}
such a function can be used as the error estimator to estimate the ROM error with a new input $\bm{J}^*$ which is excluded from the test inputs $\bm{I}=\{\bm{J}^{(1)}, \bm{J}^{(2)}, .., \bm{J}^{(s)}\}$. Based on \autoref{eq:error_interpolator} any multivariate interpolate can serve as such an error estimate. Here, we use GPR for interpolation, and the input of the interpolator is the joint input $\bm{J}$ of the ROM and the output is the estimated error.

Gaussian Process Regression \cite{williams1996gaussian}, also known as Kriging Interpolation \cite{krige1951statistical}, is a non-parametric model used for regression or interpolation. In our problem, our training dataset is the tests in the PAC validator:
\begin{equation}
    \bm{D} = (\bm{I}, \bm{e}) = \left\{\left(\bm{J}^{(i)}, e^{(i)}\right)|i=1,2,...,s\right\}
\end{equation}
According to the theory of GPR, the prediction $\hat{e}^*$ for the new joint input $\bm{J}^*$ is:
\begin{equation}
    \hat{e}^* = \bm{K}(\bm{J}^*, \bm{I})\bm{K}(\bm{I}, \bm{I})^{-1}\bm{e}
\label{eq:gpr}
\end{equation}
where $\bm{K}$ is a covariance matrix whose entries can be computed by a pre-selected covariance function $\kappa(\bm{J}^*, \bm{J}_i)$. Here we use radial-basis function (RBF), which is also called squared-exponential function or Gaussian kernel and is one of the most popular choices for the covariance function:
\begin{equation}
    \kappa(\bm{J}^*, \bm{J}^{(i)}) = \sigma^2 exp\left(-\frac{1}{2l^2}\left\|\bm{J}^*-\bm{J}^{(i)}\right\|^2\right)
\end{equation}
where the amplitude $\sigma$ and the lengthscale $l$ are the hyper-parameters which can be tuned. RBF is appropriate when the simulated function is expected to be smooth.

Besides GPR, other interpolators or approximators, such as Spline Interpolation \cite{habermann2007multidimensional}, Radial-basis-function Interpolation \cite{broomhead1988radial} and ANN can be used to construct the estimator. Here we would not test the rest methods but the readers are encouraged to try different approaches if necessary. As seen in \autoref{eq:gpr}, while fitting the GPR model, the covariance matrix $\bm{K}$ needs to be assembled and inverted. The size of $\bm{K}$ is determined by the number of test samples in $\bm{I}$. Thanks to \autoref{eq:S_PAC}, this number is always minimized.
 
\subsection{Active learning algorithm}
\label{active_learning_algorithm}
Until now, the necessary theory and components of the active learning algorithm are fully introduced. In this section we will present the whole workflow.

Firstly, we need to prepare the PAC validator as the indicator to stop the iteration. In \autoref{alg:build_PAC_validator}, the construction of the PAC validator is presented.
\begin{algorithm}
\caption{Build the PAC validator}\label{alg:build_PAC_validator}
\begin{algorithmic}[1]
\REQUIRE $\mathcal{M}$, $\sigma_\text{design}$, $\epsilon_\text{design}$
\STATE Use SPS to create $m$ IVPs for the FOM
\STATE Solve the IVPs and get the snapshots $\bm{Y}_\text{estimate}$ for estimating $\mathcal{Y}$
\STATE Construct reduced basis $\bm{V}$ using $\bm{Y}_\text{estimate}$, and estimate the initial reduced state space $\mathcal{Y}$
\STATE Loosen and trim $\mathcal{Y}$ to get $\mathcal{Y}^*$
\STATE Compute $S_\text{PAC}$ using \autoref{eq:S_PAC}
\STATE Randomly select $s$ samples ($s\geq S_\text{PAC}$) from the joint space $\mathcal{J}^*=\mathcal{Y}^*\times\mathcal{M}$, and create the inputs $\bm{I}_\text{PAC}=\left\{\bm{J}^{(1)}, \bm{J}^{(2)}, .., \bm{J}^{(s)}\right\}$ for the validator
\STATE Produce the reference outputs $\bm{\hat{O}}_\text{PAC}=\left\{\bm{\hat{y}}_\text{out}^{(1)}, \bm{\hat{y}}_\text{out}^{(2)}, .., \bm{\hat{y}}_\text{out}^{(s)}\right\}$ for the validator using the FOM solver and $\bm{V}$
\end{algorithmic}
\end{algorithm}
With the PAC validator on hands, the active learning algorithm can be designed as \autoref{alg:active_learning}.
\begin{algorithm}
\caption{Active learning for the ROM}\label{alg:active_learning}
\begin{algorithmic}[1]
\REQUIRE $\mathcal{J}^*$, $\bm{Y}_\text{estimate}$, $\bm{V}$, $\bar{\tau}^*_\text{design}$, $\Delta\bar{\tau}^*_\text{tol}$, $\text{PAC}(\tau^*,\text{ROM})$, snapshot increment $\Delta s$
\STATE Prepare a very large sample set $\bm{I}_\text{all}$ from $\mathcal{J}^*$
\STATE Randomly select $\Delta s$ samples from $\bm{I}_\text{all}$ as the initial training input samples $\bm{I}_\text{train}$ and generate the training output samples $\bm{\hat{O}}_\text{train}$ using the FOM solver
\STATE Train the ROM using $\bm{I}_\text{train}$, $\bm{\hat{O}}_\text{train}$ and $\bm{V}$
\STATE $p(\bar{\tau}^*_\text{design})=\text{PAC}(\bar{\tau}^*_\text{design};\text{ROM})$ and $\Bar{\tau}^*_\text{old}=\Bar{\tau}^*$
\STATE Compute the error snapshots $\bm{e}$ using $\bm{\hat{O}}_{\text{PAC}}$
\WHILE{$p(\bar{\tau}^*_\text{design})<1$}
    \STATE Fit the GP estimator $\mathcal{E}$ using $\bm{e}$ and $\bm{I}_{\text{PAC}}$
    \STATE Evaluate $\mathcal{E}$ on $\bm{I}_{\text{all}} \backslash \bm{I}_{\text{train}}$
    \STATE Find $\Delta s$ largest $e_i$ and their corresponding input samples $\Delta \bm{I}=\left\{\bm{J}_\text{max}^{(1)}, \bm{J}_\text{max}^{(2)}, ..., \bm{J}_\text{max}^{(\Delta s)}\right\}$
    \STATE $\bm{I}_{\text{train}} = \bm{I}_{\text{train}} \bigcup \Delta \bm{I}$
    \STATE Update $\bm{\hat{O}}_{\text{train}}$ using  the enriched $\bm{I}_{\text{train}}$ and the FOM
    \STATE Update $\bm{V}$ using $\bm{Y}_\text{estimate}\bigcup\bm{\hat{O}}_{\text{train}}$
    \STATE Train the ROM using $\bm{I}_{\text{train}}$, $\bm{\hat{O}}_{\text{train}}$ and $\bm{V}$
    \STATE $p(\bar{\tau}^*_\text{design})=\text{PAC}(\bar{\tau}^*_\text{design};\text{ROM})$ and compute $\Bar{\tau}^*$
    \STATE Update the error snapshots $\bm{e}$
    \IF{$|\Bar{\tau}^*-\Bar{\tau}^*_\text{old}|<\Delta\bar{\tau}^*_\text{tol}$}
        \STATE Break
    \ELSE
        \STATE $\Bar{\tau}^*_\text{old}=\Bar{\tau}^*$
    \ENDIF
\ENDWHILE
\end{algorithmic}
\end{algorithm}

Regarding \autoref{alg:active_learning}, we would like to clarify that the snapshots from $\bm{Y}_\text{estimate}$ are not used for the model identification step, only for the construction of the initial reduced space (i.e. initial evaluation of the projection matrix $\bm{V}$). The neural network is therefore trained exclusively with one-step trajectories. We notice the inclusion of $\bm{Y}_\text{estimate}$ does not improve the quality of the reduced model, but does considerable increase the training time for the neural network.

As seen in the algorithm, the reduced basis $\bm{V}$ is re-computed during the iterations. This theoretically should influence the estimate of the reduced-state space and would require re-building the PAC validator using the FOM solver. However, in practice, the initial basis $\bm{V}$ produced by $\bm{Y}_\text{estimate}$ should already have relatively low projection error. The refined reduced basis only slightly changes the boundary of the reduced-state space. Since we use an expansion ratio $\beta$ to loose the estimated boundary, we can still use $\mathcal{Y}$ as the space for sampling the reduced state.

As mentioned on the end of \autoref{sec:ROM_validator}, in \autoref{alg:active_learning} we see another convergence check beside checking if $p(\bar{\tau}^*_\text{design})\geq 1$. Additionally, we check the improvement between two iterations. $1-\Bar{\tau}^*$ is the best ROM accuracy in the current iteration that we can have at the given confidence level $\eta_\text{design}$. And $1-\Bar{\tau}^*_\text{old}$ is the best ROM accuracy in the last iteration. If the difference between them is already smaller than a predefined value $\Delta\bar{\tau}^*_\text{tol}$, the algorithm will be considered to be in a low-efficient status and should be stopped to save computational resource. Since there are many factors influencing the ROM accuracy other than the amount of training data, e.g. ROM structure, ROM dimension, the way of collecting training data, etc., other possible modification should be considered and adopted to improve the ROM accuracy. 

\section{Numerical examples}
\label{sec:numerical_examples}
In this section, two numerical examples will be used to evaluate the performance of the proposed algorithm. The full-order problems have different complexity. The first problem is a 2-D linear heat problem, while the second one is a 3-D nonlinear thermal radiation problem. The conventional machine-learning-based MOR and the active-learning-based MOR is applied to both problems. Their performance is analyzed and compared.

\subsection{2-D linear thermal system}
The governing equation of the test system is:
\begin{align}
    \begin{split}
        &\frac{\partial T(x, y, t)}{\partial t} = k_x \frac{\partial^2 T(x, y, t)}{\partial x^2} + k_y \frac{\partial^2 T(x, y, t)}{\partial y^2}\\
        &T(0, y, t) = T_{\text{left}}, \qquad T(10, y, t) = T_{\text{right}}\\
        &T(x, -10, t) = T_{\text{down}}, \qquad T(x, 10, t) = T_{\text{up}}\\
        &T(x, y, 0) = 20
    \end{split}
    \label{eq:2d_heat_equation}
\end{align}
where $x\in[0, 10], y\in[-10, 0], t\in[0, 2]$ and the static parameters are given by $k_x=k_y=1$. And the controllable parameters $T_{\text{left}}, T_{\text{right}}, T_{\text{down}}$ and $T_{\text{up}}$ ($^\circ C$) have the same feasible range $[20 , 1000]$. Thus, for this problem $\mathcal{M} = [20, 1000]\times[20, 1000]\times[20, 1000]\times[20, 1000]$. The full order model is discretized by finite difference method. The square region is divided into $50\times 50=2500$ cells. We consider 100 time steps uniformly distributed.

For the active learning algorithm, we first need to estimate the reduced-state space. Thus, $m=20$ samples are taken from $\mathcal{M}$. The 20 constant-input IVPs are constructed with the sampled parameters. $\bm{Y}_\text{estimate}$ has totally 2,020 state vectors. The initial reduced space is constructed by 20 POD basis using $\bm{Y}_\text{estimate}$. The low-dimensional projection of those state vectors are then used to perform the space estimate. After estimating, loosening and trimming, we create the joint space $\mathcal{J}^*=\mathcal{Y}^*\times\mathcal{M}$ and use random sampling to get 40,000 joint samples for $\bm{I}_\text{all}$.

The ROM will be based on the following architecture of the EENN. The MLP in the EENN has four layers, and each layer has [20, 80, 80, 20] neurons. We define $\epsilon=3\%$ and $\sigma=1\%$ for the validation step. Based on \autoref{eq:S_PAC}, we know 2,944 samples are needed by the PAC validator. The confidence level that will be investigated is $\eta_\text{design}=1-3\%=97\%$. The snapshots increment $\Delta s$ is selected to be 500, which means another 500 snapshots will be added into the training set between the iterations. And we pick $\bar{\tau}^*_\text{design}=1\%$ and $\Delta\bar{\tau}^*_\text{tol}=0.01\%$. In conclusion, we are looking for a ROM which has $97\%$ confidence of being $99\%$ accurate in each prediction.

As seen in \autoref{tab:AL_convergence_TS}, the ROM does not converge to the prescribed accuracy after 10 iterations. However, it meets the other stopping criteria $\Delta\bar{\tau}^*_\text{tol}\leq0.01\%$. This means, we consider that a ROM with the current settings cannot be better than this upper limit ($1-\Bar{\tau}^*=98.17\%$) even given with more training data.
\begin{table}[h!]
\begin{center}
    \begin{tabular}{ *{4}{c} }
    \hline
    Iteration & Num. samples & $1-\Bar{\tau}^*$ & $p(\bar{\tau}^*_\text{design})$ \\
     \hline
     1 & 500 & $94.70\%$ & $6.05\%$\\
     \hline
     2 & 1,000 & $95.60\%$ & $42.63\%$\\
     \hline
     3 & 1,500 & $97.00\%$ & $88.04\%$\\
     \hline
     4 & 2,000 & $97.50\%$ & $95.35\%$\\
     \hline
     5 & 2,500 & $97.60\%$ & $95.79\%$\\
     \hline
     6 & 3,000 & $97.47\%$ & $96.06\%$\\
     \hline
     7 & 3,500 & $97.80\%$ & $96.50\%$\\
     \hline
     8 & 4,000 & $98.13\%$ & $96.67\%$\\
     \hline
     9 & 4,500 & $98.17\%$ & $96.77\%$\\
     \hline
     10 & 5,000 & $98.17\%$ & $96.64\%$\\
     \hline
    \end{tabular}
\end{center}
\caption{PAC evaluation: AL-MOR}
\label{tab:AL_convergence_TS}
\end{table}

For the conventional workflow, we use DPS to create 50 IVPs, and they produce 50 solution trajectories. 101 outputs from each IVP can build 100 input-output samples, therefore, we totally have 5,000 training samples. To construct the POD space, besides the sampled 5,000 snapshots, we also include $\bm{Y}_\text{estimate}$.

The same PAC validator is used to evaluate the ROM produced by the conventional workflow and the result is given in \autoref{tab:DPL_convergence_TS}.
\begin{table}[h!]
    \begin{center}
        \begin{tabular}{ *{3}{c} }
        \hline
        Num. samples & $1-\Bar{\tau}^*$ & $p(\bar{\tau}^*_\text{design})$ \\
         \hline
         5000 & $89.63\%$ & $47.96\%$ \\ 
         \hline
        \end{tabular}
    \end{center}
    \caption{PAC evaluation: conventional MOR}
    \label{tab:DPL_convergence_TS}
\end{table}

It is observed that with the same amount of training samples, the two performance indicators $\Bar{\tau}^*$ and $p(\bar{\tau}^*_\text{design})$ of the conventional machine learning ROM are both much lower than the ROM trained in the active learning algorithm. Based on the PAC theory, the conventional ROM (ML-ROM) is not expected to have more reliable performance compared with the active-learning-based ROM (AL-ROM).

\begin{figure}[!htbp]
    \centering
    \includegraphics[width=0.49\textwidth]{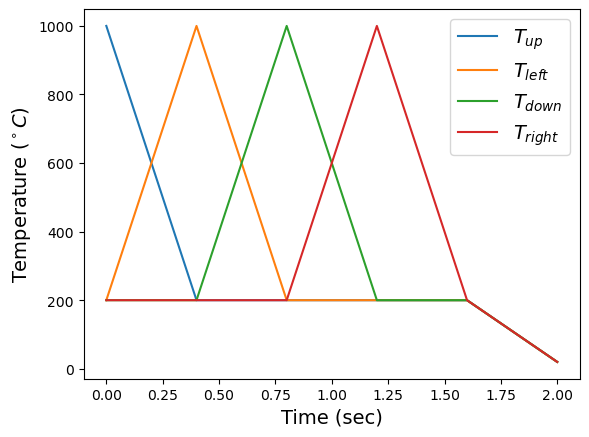}
    \caption{Time-dependent inputs to the 2-D thermal problem for test 1.}
    \label{fig:heat_eq_inputs}
\end{figure}

\begin{figure}[!htbp]
    \centering
    \includegraphics[width=0.6\textwidth]{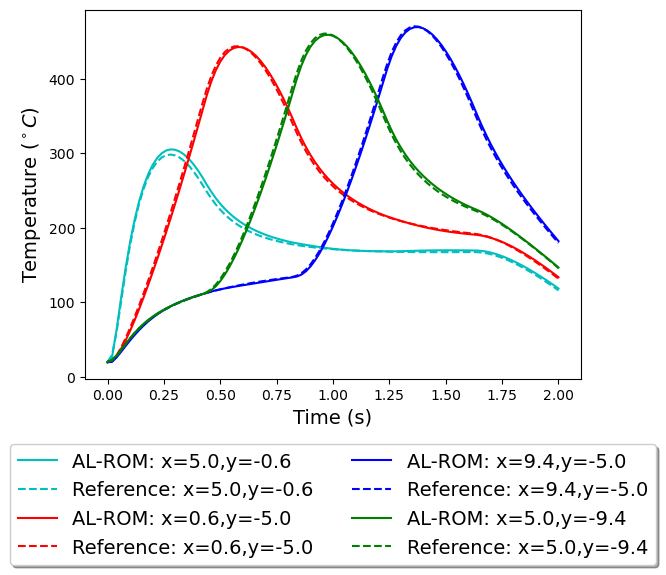}
    \caption{
        Comparison between FOM and ROM trajectories in the test 1. The ROM is produced by the active learning workflow.
        }
    \label{fig:heat_eq_comparison_1_AL}
\end{figure}

\begin{figure}[!htbp]
    \centering
    \includegraphics[width=0.6\textwidth]{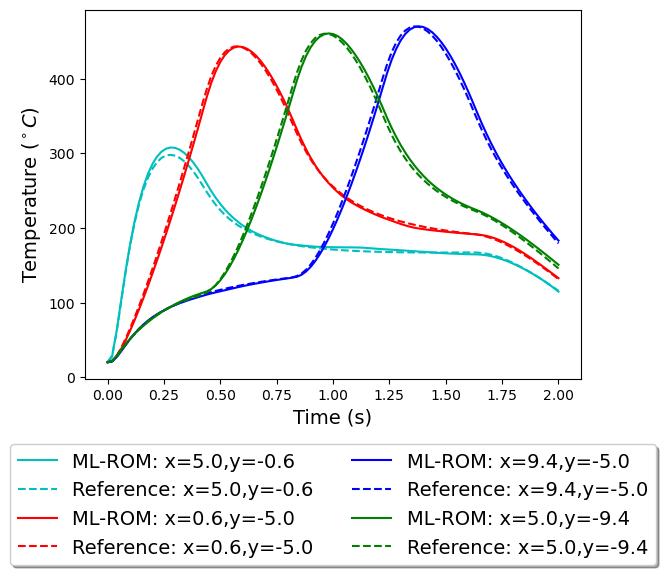}
    \caption{
        Comparison between FOM and ROM trajectories in the test 1. The ROM is produced by the conventional workflow.
        }
    \label{fig:heat_eq_comparison_1_DPL}
\end{figure}

\begin{figure}[!htbp]
    \centering
    \includegraphics[width=0.6\textwidth]{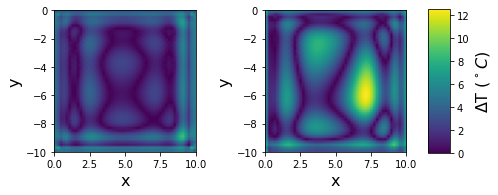}
    \caption{
        Comparison between two ROMs' error (absolute) fields at $t=2s$ in the test 1. Left: by active learning workflow. Right: by conventional machine learning workflow.
        }
    \label{fig:heat_eq_error_field_1}
\end{figure}

To further validate the ROMs, we designed two additional test cases. In test 1, 4 boundary temperatures are described by 4 different time-dependent functions respectively. The function curves are given in \autoref{fig:heat_eq_inputs}. And we pick 4 different elements as the observation positions. In \autoref{fig:heat_eq_comparison_1_AL} and \autoref{fig:heat_eq_comparison_1_DPL} we present the predicted trajectories by the compared ROMs. In the trajectory figures, it is observed that both ROMs' prediction is very close to the reference and it is hard to decide which prediction is outperforming. However, if we look at \autoref{fig:heat_eq_error_field_1}, where their error fields at the final time step is presented. To compute the error field, we first lift the reduced solution to the full space using $\bm{V}$, and calculate the absolute error between the lifted solution and the FOM solution. We can see the error field of the AL-ROM is overall better than its competitor's.

\begin{figure}[!htbp]
    \centering
    \includegraphics[width=0.48\textwidth]{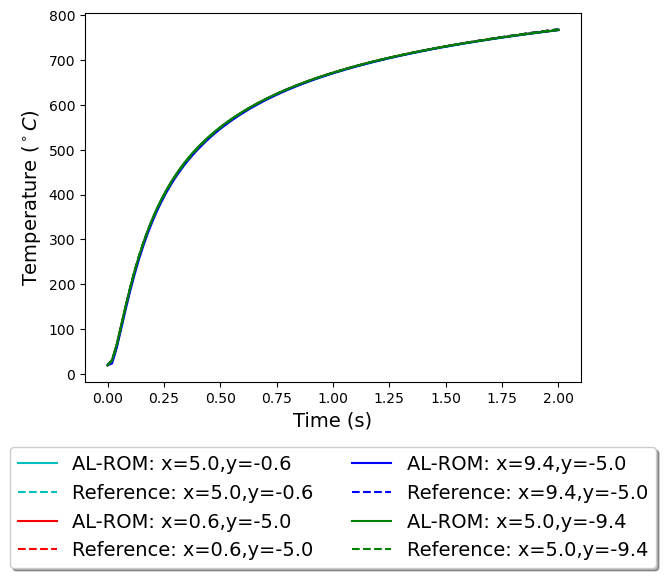}
    \caption{
        Comparison between FOM and ROM trajectories in the test 2. The ROM is produced by the active learning workflow.
        }
    \label{fig:heat_eq_comparison_2_AL}
\end{figure}

\begin{figure}[!htbp]
    \centering
    \includegraphics[width=0.6\textwidth]{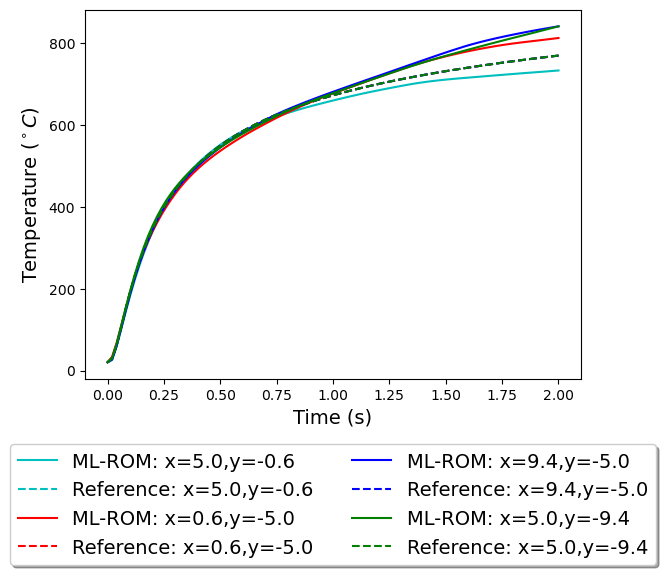}
    \caption{
        Comparison between FOM and ROM trajectories in the test 2. The ROM is produced by the conventional workflow.
        }
    \label{fig:heat_eq_comparison_2_DPL}
\end{figure}

\begin{figure}[!htbp]
    \centering
    \includegraphics[width=0.6\textwidth]{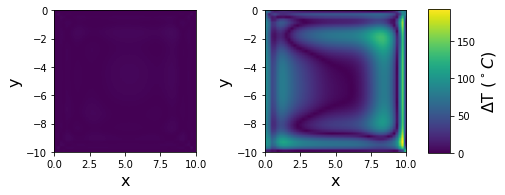}
    \caption{
        Comparison between two ROMs' error (absolute) fields at $t=2s$ in the test 2. Left: by active learning workflow. Right: by conventional machine learning workflow.
        }
    \label{fig:heat_eq_error_field_2}
\end{figure}

\begin{figure}[!htbp]
    \centering
    \includegraphics[width=0.55\textwidth]{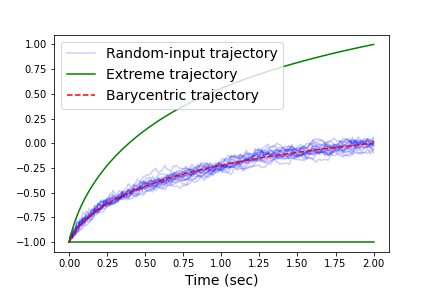}
    \caption{The trajectories of the $1^\text{st}$ POD coefficient formed by: random-input samples (blue), extreme-input samples (green), and medium-input samples (dashed red). Here the extreme inputs are $[1000, 1000, 1000, 1000]$ for the upper green curve and $[20, 20, 20, 20]$ for the lower green curve. The medium input is $[510, 510, 510, 510]$.}
    \label{fig:trajectory_comparison}
\end{figure}

In test 2, 4 boundary temperatures are assigned with the same constant boundary temperatures $T_{\text{left}}=T_{\text{right}}=T_{\text{down}}=T_{\text{up}}=1000^\circ C$. The trajectories are presented in \autoref{fig:heat_eq_comparison_2_AL} and \autoref{fig:heat_eq_comparison_2_DPL}. And the error fields are shown in \autoref{fig:heat_eq_error_field_2}. In test 2, it is observed that the trajectories predicted by the conventional machine learning ROM have large difference from the reference, while the AL-ROM's prediction can still match the reference. And if we check the error fields at the last time step, the AL-ROM's advantage is even more remarkable. Here a possible reason can be explained by \autoref{fig:trajectory_comparison}.

In \autoref{fig:trajectory_comparison}, to simplify the problem, we only consider the first POD coefficient of $\bm{y}_r$. We create an IVP whose inputs are $[1000, 1000, 1000, 1000]$. Knowing the physics of the original problem, we know that this is the parameter combination that would generate the largest temperature increment of the whole system. Similarly, another trajectory is produced by giving the parameter combination $[20, 20, 20, 20]$ to the FOM. These two trajectories are called "extreme trajectories". Since our full order problem is a continuous problem, it is reasonable to assume that any trajectory produced by any parameter combination in $\mathcal{M}$ should be between these two green boundaries. Then we pick a parameter combination $[510, 510, 510, 510]$, which can be considered as the barycenter of $\mathcal{M}$, and the corresponding trajectory is drawn in the red dashed line and called "barycentric trajectory". It is observed that the trajectories produced by the IVPs in the conventional workflow form a thin band whose center is approximately the barycentric trajectory, and this phenomenon is also observed and explained in \cite{astrom1970chapter}. This means assigning randomized parameter combinations to the IVP does not produce a wide distribution of the solution trajectory. A result of this narrow distribution is that the ROM will perform badly beyond the barycentric trajectory, the accuracy can be very bad. This is reflected in the test 2. Recalling the sampling in the active learning workflow, different $\bm{y}_r$ are randomly sampled in the estimated space $\mathcal{Y}$ and used as the initial states for the one-step prediction. This distributes the training samples in a wider region than the thin band and the ROM trained by such dataset should also be more reliable facing different inputs.

Compared with the ANN-ROM trained by the data collected by the conventional sampling method, the ROM constructed by our method outperforms with the same amount of training snapshots. This advantage can not only be observed in the toy example, but also in a more realistic setting.

\subsection{3-D vacuum furnace model}
\label{sec:3d_vf_model}
\begin{figure}[!htbp]
    \centering
    \includegraphics[width=0.35\textwidth]{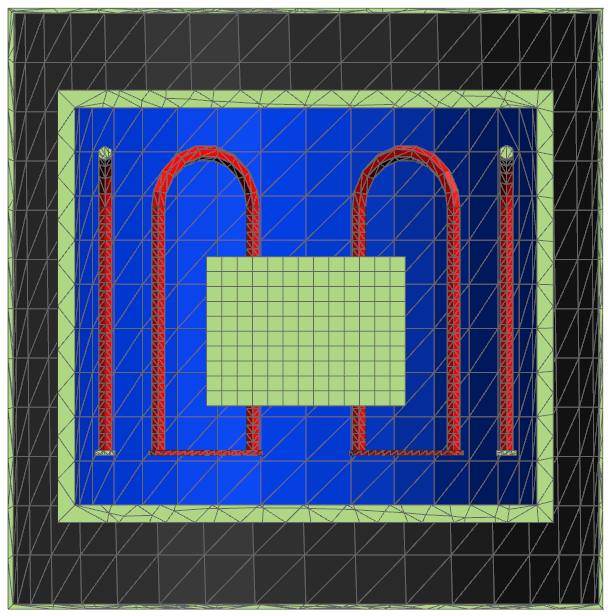}
    \caption{The cross-section of the 3-D simulation model of the vacuum radiation furnace.}
    \label{fig:vf_model}
\end{figure}
In \autoref{fig:vf_model}, we present the 3-D simulation model of a vacuum radiation furnace \cite{hao20083}. The model was created and simulated using \href{https://www.plm.automation.siemens.com/global/en/products/simcenter/}{NX Simcenter 3D}. The original FEM model has 20,209 elements. The dark grey part represents the outer case of the furnace. The red parts are heaters for which there are 6 in total. The blue component between the heaters and the case is the protecting shell. The cubic area at the center is the workzone where material can be placed. Since the internal space of the furnace is vacuum, during the industrial process, the material placed in the furnace will only be heated up by the thermal radiation from the heaters. The discrete governing equation of the system can be written as:
\begin{equation}
    \bm{E}\frac{\partial\bm{T}}{\partial t}=\bm{K}\bm{T}+\bm{R}\bm{T}^4+\bm{B}\bm{u}
    \label{eq:vf_governing_eq}
\end{equation}
where $\bm{E}$ is thermal capacity matrix, $\bm{K}$ is thermal conductivity matrix, $\bm{R}$ is radiation matrix, and $\bm{B}$ is heat-load-distribution matrix. The state (temperature) vector of this system is $\bm{T}\in\mathbb{R}^N$ where $N=20,209$. We consider there are 7 controllable parameters in the system: 6 heating power of the heaters ($kW/m^3$) and the convection coefficient ($W/(m^2K)$) of the outer surface of the case. Therefore, we have the load vector $\bm{u}\in\mathbb{R}^{N_u}$ with $N_u=7$. The parameter space is predefined to be:
\begin{align}
    \begin{split}
        \mathcal{M}=&\left[0, 10^4\right]\times\left[0, 10^4\right]\times\left[0, 10^4\right]\times\left[0, 10^4\right]\\
        &\times\left[0, 10^4\right]\times\left[0, 10^4\right]\times\left[5, 100\right]
    \end{split}
    \label{eq:M_for_vf_model}
\end{align}
Based on physical consideration, the time grid is from 0 to 45,000 seconds with 450 time steps.

For the active learning algorithm, $m=20$ samples are taken from $\mathcal{M}$ to estimate the boundaries of $\mathcal{Y}$. We also include two special parameter configurations:
\begin{align}
    \begin{split}
        &\bm{\mu}_{T_\text{min}} = \left[0, 0, 0, 0, 0, 0, 100\right]\\
        &\bm{\mu}_{T_\text{min}} = \left[10^4, 10^4, 10^4, 10^4, 10^4, 10^4, 5\right]
    \end{split}
    \label{eq:extrem_mu_VF_model}
\end{align}
The initial reduced space is constructed by $N_r=20$ POD basis. The low-dimensional projection of those state vectors are then used to perform the space estimate. Then we create the joint space $\mathcal{J}^*=\mathcal{Y}^*\times\mathcal{M}$ and use random sampling to get 40,000 joint samples for $\bm{I}_\text{all}$.

We define $\epsilon=3\%$ and $\sigma=1\%$. And 2,944 samples are collected for the PAC validator. The investigated confidence level is $\eta_\text{design}=97\%$. $\Delta s$ is selected to be 2,500. We pick $\bar{\tau}^*_\text{design}=1.5\%$ and $\Delta\bar{\tau}^*_\text{tol}=0.01\%$. In conclusion, we are looking for a ROM which has $97\%$ confidence of being $98.5\%$ accurate in each prediction.

The AL-ROM's convergence is presented in \autoref{tab:AL_convergence_VF}. The final ROM accuracy is $97.99\%$. For the conventional approach, the final ROM has only $84.42\%$ accuracy at the investigated confidence level. When the AL-ROM is accepted, it uses 20,000 $\bm{J}$ for training. Accordingly, we construct a ML-ROM using DPS with the same number of snapshot for ROM identification. And to be fair, we also include $\bm{Y}_\text{estimate}$ into the snapshot matrix while constructing the POD space.
\begin{table}[!htbp]
\begin{center}
    \begin{tabular}{ *{4}{c} }
    \hline
    Iteration & Num. samples & $1-\Bar{\tau}^*$ & $p(\bar{\tau}^*_\text{design})$ \\
     \hline
     1 & 2,500 & $94.97\%$ & $57.92\%$\\
     \hline
     2 & 5,000 & $96.81\%$ & $96.80\%$\\
     \hline
     3 & 7,500 & $97.11\%$ & $98.36\%$\\
     \hline
     4 & 10,000 & $97.49\%$ & $98.48\%$\\
     \hline
     5 & 12,500 & $97.83\%$ & $98.68\%$\\
     \hline
     6 & 15,000 & $97.93\%$ & $98.84\%$\\
     \hline
     7 & 17,500 & $97.99\%$ & $98.88\%$\\
     \hline
     8 & 20,000 & $97.99\%$ & $98.88\%$\\
     \hline
    \end{tabular}
\end{center}
\caption{PAC evaluation: AL-MOR}
\label{tab:AL_convergence_VF}
\end{table}

\begin{figure}[!htbp]
    \centering
    \includegraphics[width=0.55\textwidth]{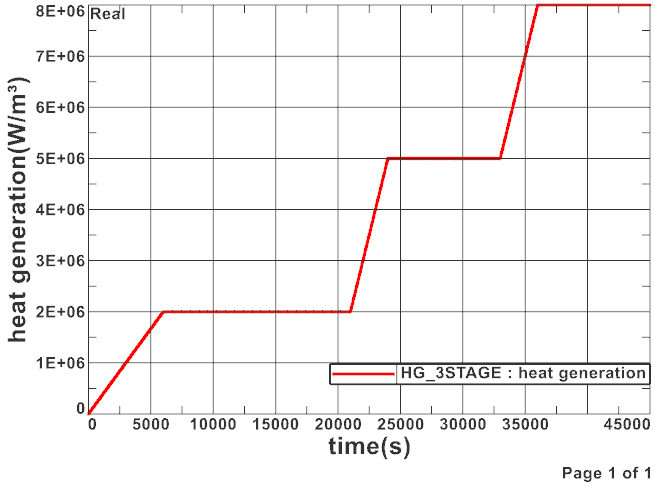}
    \caption{The heat generation for the first test case. The strategy is applied to all 6 heaters.}
    \label{fig:hl_3stage}
\end{figure}

\begin{table}[h!]
    \begin{center}
        \begin{tabular}{ *{3}{c} }
        \hline
        Num. samples & $1-\Bar{\tau}^*$ & $p(\bar{\tau}^*_\text{design})$ \\
         \hline
         20,000 & $84.42\%$ & $1.92\%$ \\ 
         \hline
        \end{tabular}
    \end{center}
    \caption{PAC evaluation: conventional MOR}
    \label{tab:DPL_convergence_VF}
\end{table}
Two test scenarios are employed to show the ROMs' performance in relatively realistic conditions. In the first test, a 3-stage heating profile \autoref{fig:hl_3stage} is used and the convection coefficient is set to be $50W/(m^2K)$.
In this test case, we monitor the temperature trajectories on a heater and in the workzone.
\begin{figure}[htbp]
    \centering
    \subfloat[AL-ROM]{\includegraphics[width=0.48\textwidth]{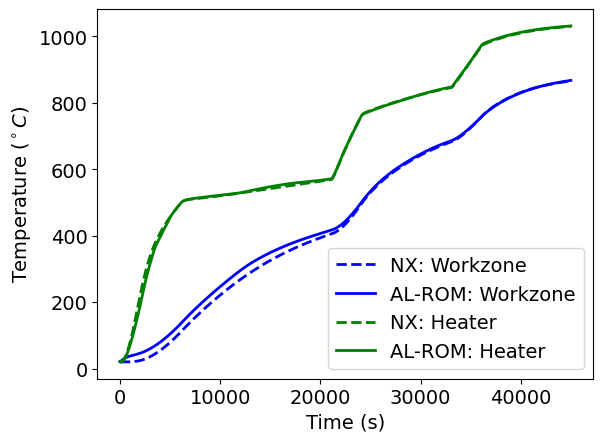}\label{fig:vf_3stage_al}}
    \hfill
    \subfloat[ML-ROM]{\includegraphics[width=0.48\textwidth]{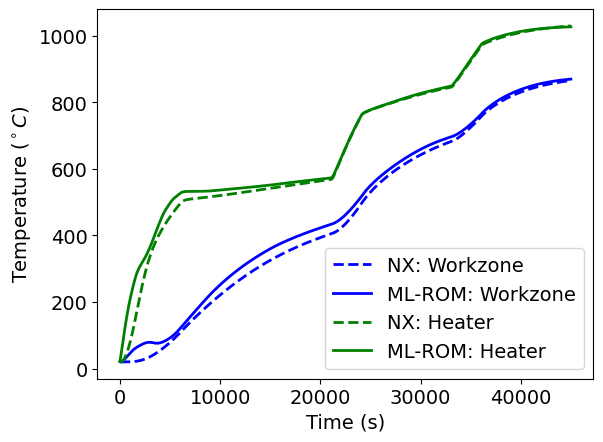}\label{fig:vf_3stage_ml}}
    \caption{Comparison between the NX solution and the ROMs' prediction for the 3-stage heating profile.}
    \label{fig:vf_3stage_test}
\end{figure}
And to show the error for the whole temperature field, we compute the absolute errors for all 20,209 elemental temperature individually and 4 statistical measures, mean, standard deviation (std.), minimum and maximum, to reflect the error distribution at the final time point. The results are shown in \autoref{tab:3stage_error_field}.
\begin{table}[!htbp]
    \begin{center}
        \begin{tabular}{ *{3}{c} }
         \hline
           & AL-ROM & ML-ROM \\
         \hline
         Mean ($^\circ C$)& 3.21 & 24.78 \\
         \hline
         Std. ($(^\circ C)^2$)& 2.91 & 19.97 \\
         \hline
         Min. ($^\circ C$) & 2.00E-5 & 1.64E-4 \\
         \hline
         Max. ($^\circ C$)& 11.60 & 195.18 \\
         \hline
        \end{tabular}
    \end{center}
    \caption{Statistical analysis for the error field ($\Delta T$) at $t=45000s$ for the 3-stage heating profile.}
    \label{tab:3stage_error_field}
\end{table}

\begin{figure}[!htbp]
    \centering
    \includegraphics[width=0.4\textwidth]{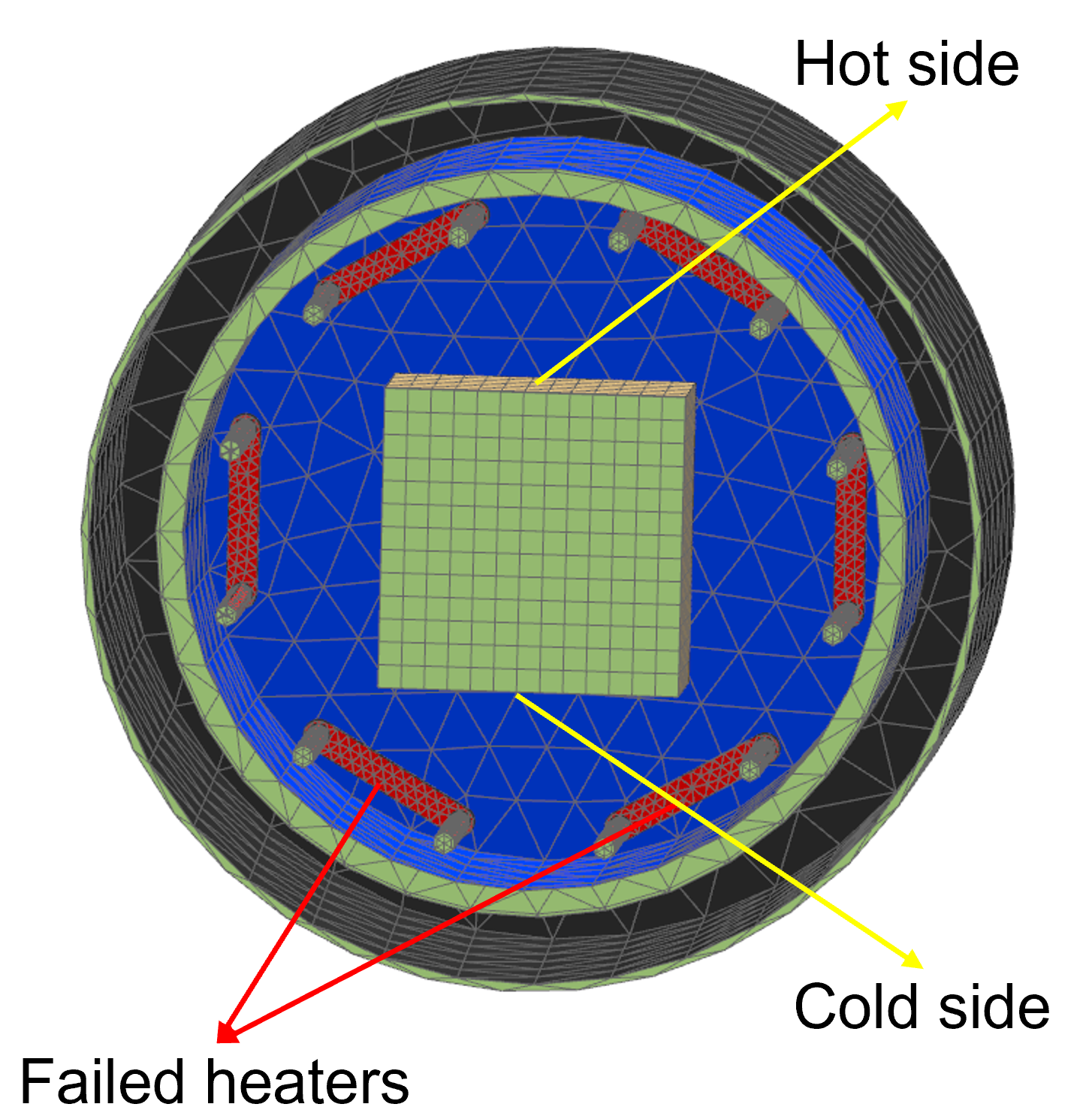}
    \caption{The failure case of the vacuum furnace model.}
    \label{fig:vf_failure_case}
\end{figure}

In another test case, we use the same heating profile for 4 heaters and turn off 2 heaters (\autoref{fig:vf_failure_case}). This simulates a scenario where 2 heaters have failed during industrial process.
And the existence of the failed heaters will cause non-uniform temperature field in the workzone. We call them cold side and hot side respectively. In this test, we monitor the temperature trajectories of the cold side and the hot side using the ROMs.
\begin{figure}[htbp]
    \centering
    \subfloat[AL-ROM]{\includegraphics[width=0.48\textwidth]{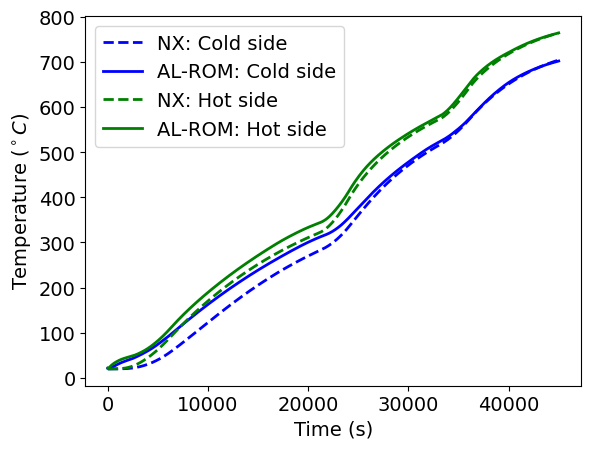}\label{fig:vf_failure_al}}
    \hfill
    \subfloat[ML-ROM]{\includegraphics[width=0.48\textwidth]{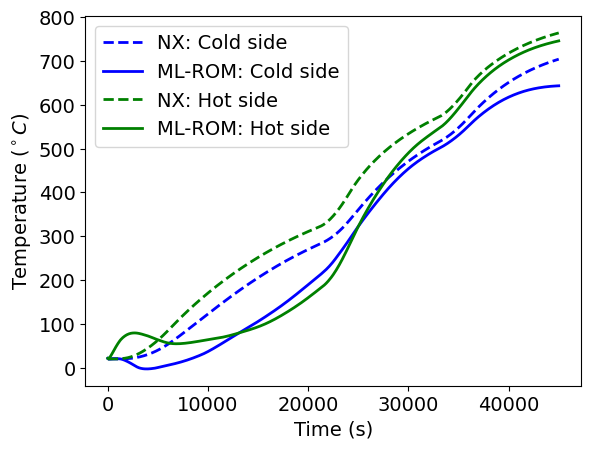}\label{fig:vf_failure_ml}}
    \caption{Comparison between NX solution and ROMs' prediction for the heating profile with failure.}
    \label{fig:vf_failure_test}
\end{figure}
\begin{table}[!htbp]
    \begin{center}
        \begin{tabular}{ *{3}{c} }
         \hline
           & AL-ROM & ML-ROM \\
         \hline
         Mean ($^\circ C$)& 1.52 & 63.51 \\
         \hline
         Std. ($(^\circ C)^2$)& 0.81 & 59.31 \\
         \hline
         Min. ($^\circ C$) & 8.47E-5 & 2.86E-3 \\
         \hline
         Max. ($^\circ C$)& 6.17 & 401.68 \\
         \hline
        \end{tabular}
    \end{center}
    \caption{Statistical analysis for the error field ($\Delta T$) at $t=45000s$ for the heating profile with failure.}
    \label{tab:failed_error_field}
\end{table}
The corresponding comparison between ROM-predicted trajectories and FOM-predicted trajectories in two test cases are given in \autoref{fig:vf_failure_test} and \autoref{tab:failed_error_field}.

As observed in both trajectory comparison in \autoref{fig:vf_3stage_test} and \autoref{fig:vf_failure_test}, the ROM constructed by the active-learning algorithm have better agreement to the reference result. Moreover, in \autoref{tab:3stage_error_field} and \autoref{tab:failed_error_field}, all statistical measurement of the AL-ROM's error fields are better than the competitor's. These facts tell us the AL-ROM has generally better performance in this non-linear thermal system.

\section{Conclusion}
\label{sec:conclusion}
In this work, we proposed a new approach to constructing the ANN-based ROMs. The approach is based on the principle of active learning. To achieve this goal, a new snapshot collection strategy and a new ROM validator is proposed. With the new snapshot collection method, we will create a joint space consisting of a predefined parameter space and an estimated reduced-state space. The joint samples randomly taken from such a space will contain the initial states and the system parameters for the one-step snapshots. Such snapshots are expected to have better distribution in the state space compared with the snapshots obtained in conventional way. Besides, a new validator is constructed based on PAC theory. Different from the conventional validation where case-dependent tests are employed, the PAC validator evaluates ROMs in a statistical way. The confidence and the accuracy of ROMs is computed and used to describe the general performance of ROMs. And during the validation, error snapshots will be computed for training a data-driven error estimator based on GP. The error estimator is employed to instruct the collection of snapshots.

We use two numerical models to test the different approaches. One is a linear-thermal-conduction problem, and the other one uses the simulation model of a vacuum radiation furnace. In both problems we use the proposed algorithm to construct AL-ROMs and use conventional approach for the competitive ML-ROMs.

In all the tests we find that with the same amount of training data, the active-learning approach works generally better. Collecting training samples from estimated joint space provides us wider exploration not only for system parameters, but also for reduced states. Besides, the feasibility of using PAC score as a performance indicator is also proven in the tests. An ROM with higher PAC score can be considered to be more accurate and stable also in case-dependent validation.

The ROMs constructed in such way can predict with complex input parameters. This will allow us to use MOR to achieve real-time simulation for more realistic FOMs in industry.

\bibliographystyle{unsrt}  
\bibliography{references}

\end{document}